\documentclass[10pt,journal,compsoc]{IEEEtran}
\ifCLASSOPTIONcompsoc
  \usepackage[nocompress]{cite}
\else
  \usepackage{cite}
\fi
\ifCLASSINFOpdf
  \usepackage[pdftex]{graphicx}
  \graphicspath{{../pdf/}{../jpeg/}}
  \DeclareGraphicsExtensions{.pdf,.jpeg,.png}
\else
  \usepackage[dvips]{graphicx}
  \graphicspath{{../eps/}}
  \DeclareGraphicsExtensions{.eps}
\fi
\usepackage{amsmath}
\usepackage{array}

\usepackage[pagebackref=true,breaklinks=true,colorlinks]{hyperref}
\usepackage{url}
\usepackage{bm}

\def\foo#1{\xfoo#1\relax^\relax\valign}
\def\xfoo#1^#2\relax#3\valign{%
\mathbf{#1}\ifx\valign#2\valign\else^{\mathbf{#2}}\fi}

\def\eg{\emph{e.g}.} 
\def\ie{\emph{i.e}.} 
 
 \def\vs{\emph{vs}.}
 
\def\etal{\emph{et al}.}

\newcolumntype{L}[1]{>{\raggedright\let\newline\\\arraybackslash\hspace{0pt}}m{#1}}
\newcolumntype{R}[1]{>{\raggedleft\let\newline\\\arraybackslash\hspace{0pt}}m{#1}}
\newcolumntype{C}[1]{>{\centering\let\newline\\\arraybackslash\hspace{0pt}}m{#1}}
\newcolumntype{x}{>\small c}

\usepackage{amssymb}
\usepackage{color}

\begin{document}
% \title{Object Detection in Videos by Short and Long Range Object Linking}
\title{Object Detection in Videos by High Quality Object Linking}

\author{Peng~Tang,
        Chunyu~Wang,
        Xinggang~Wang,
        Wenyu~Liu,
        Wenjun~Zeng,
        and~Jingdong~Wang% <-this % stops a space
\IEEEcompsocitemizethanks{\IEEEcompsocthanksitem P. Tang, X. Wang, and W. Liu are with the School of Electronic Information and Communications, Huazhong University of Science and Technology, Wuhan, China.\protect\\
E-mail: \{pengtang, xgwang, liuwy\}@hust.edu.cn \protect\\
 C. Wang, W. Zeng, and J. Wang are with Microsoft Research, Beijing, China.\protect\\
E-mail: \{chnuwa, wezeng, jingdw\}@microsoft.com \protect\\
Corresponding authors: Jingdong Wang and Wenyu Liu
}% <-this % stops an unwanted space
% \thanks{Manuscript received April 19, 2005; revised August 26, 2015.}
}


\markboth{Journal of \LaTeX\ Class Files}%
{Tang \MakeLowercase{\textit{et al.}}: Object Detection in Videos by High Quality Object Linking}



\IEEEtitleabstractindextext{%
\begin{abstract}
Compared with object detection in static images,
object detection in videos is more challenging
due to degraded image qualities.
An effective way to address this problem is
to exploit temporal contexts
by linking the same object across video to form tubelets
and aggregating classification scores in the tubelets.
In this paper,
we focus on obtaining high quality object linking results
for better classification.
Unlike previous methods that link objects
by checking boxes between neighboring frames,
we propose to link in the same frame.
To achieve this goal,
we extend prior methods in following aspects:
(1)~a cuboid proposal network
that extracts spatio-temporal candidate cuboids
which bound the movement of objects;
(2)~a short tubelet detection network
that detects short tubelets
in short video segments;
(3)~a short tubelet linking algorithm
that links temporally-overlapping short tubelets
to form long tubelets.
Experiments on the ImageNet VID dataset show
that our method outperforms
both the static image detector
and the previous state of the art.
In particular,
our method improves results by $8.8\%$ over
the static image detector for fast moving objects.
\end{abstract}

}

\maketitle

\IEEEdisplaynontitleabstractindextext

% \fi
\IEEEpeerreviewmaketitle

\IEEEraisesectionheading{\section{Introduction}
\label{sec:introduction}}

\IEEEPARstart{D}{etecting} objects in static images \cite{girshick2015fast,girshick2014rich,liu2016ssd,redmon2016you,ren2015faster,tang2017multiple,zhang2018single} has achieved
significant progress
due to the emergence of deep convolutional neural networks (CNNs) \cite{he2016deep,krizhevsky2012imagenet,lecun1998gradient,simonyan2015very}.
However,
object detection in videos brings additional challenges
due to degraded image qualities,
\eg\ motion blur and video defocus,
leading to unstable classifications
for the same object across video.
Therefore, many research efforts have been allocated to video object detection
by exploiting temporal contexts \cite{han2016seq,feichtenhofer2017detect,kang2016object,kang2016t,kang2017object,zhu2016deep,zhu2017flow,wang2015visual,bertasius2018object,xiao2018video},
especially after the introduction of the ImageNet video object detection (VID) challenge.

Many previous methods exploit temporal contexts
by linking the same object across video to form tubelets
and aggregating classification scores in the tubelets \cite{han2016seq,kang2016object,kang2016t,feichtenhofer2017detect}.
They first use static image detectors to detect objects in each frame,
and then link these detected objects
by checking object boxes between neighboring frames,
according to the spatial overlap between object boxes in different frames \cite{han2016seq}
or predicting object movements between neighboring frames \cite{kang2016object,kang2016t,kang2017object,feichtenhofer2017detect}.
Very promising results are obtained by these methods.

However, the same object changes its locations and appearances in neighboring frames
due to object motion,
which may make the spatial overlap between boxes of the same object
in neighboring frames not sufficient enough
or the predicted object movements not accurate enough.
This influences the quality of object linking,
especially for fast moving objects.
By contrast,
in the same frame,
it is obvious
that two boxes
correspond to the same object
if they have sufficient spatial overlaps.
Inspired by these facts,
we propose to link objects in the same frame
instead of neighboring frames
for high quality object linking.

\begin{figure}
\centering
\footnotesize
\includegraphics[width=\linewidth]{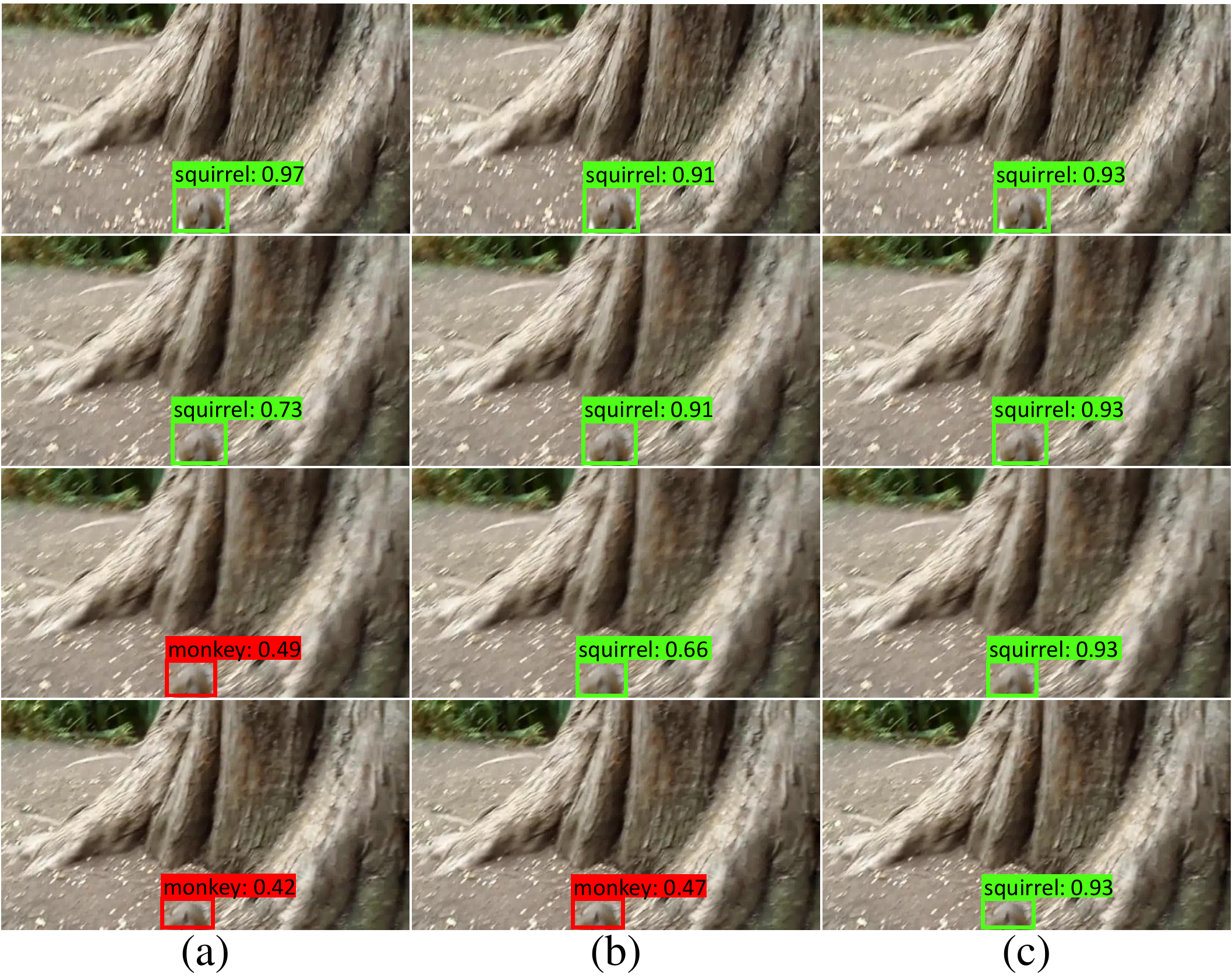}
\caption{
(a) Static image detection:
only the detections in the first two frames are correct.
(b) Short tubelet detection:
the detection in the third frame becomes correct
due to object linking between the second and third frame.
(c) Short tubelet linking:
The detections in all the frames are correct
due to short tubelet linking.
Here the short video segment length is 2.
For each frame, we only show the top-scoring box,
where green/red boxes correspond to success/failure examples.
}
\label{fig:teaser}
%\vspace{-.4cm}
\end{figure}

In our method,
a long video is first divided into
some temporally-overlapping short video segments.
For each short video segment,
we extract a set of cuboid proposals,
\ie\ spatio-temporal candidate cuboids
which bound the movement of objects,
by extending the region proposal network
for static images \cite{ren2015faster}
to a cuboid proposal network
for short video segments.
The objects across frames lying in a cuboid
are regarded as the same object.
{The main benefit from cuboid proposal
is to enable object linking in the same frame
and it alone yields minor detection performance improvement.}

For each cuboid proposal,
we adapt the Fast R-CNN~\cite{girshick2015fast}
to detect short tubelets.
More precisely,
we compute the precise box locations
and classification scores for each frame separately,
forming a short tubelet representing the linked object boxes
in the short video segment.
We compute the classification score of the tubelet,
by aggregating the classification scores of the boxes across frames.
In addition, to remove spatially redundant short tubelets,
we extend the standard non-maximum suppression (NMS)
with a tubelet overlap measurement,
which prevents tubelets from breaking
that may happen in frame-wise NMS.
Considering short range temporal contexts
by short tubelets
benefits detection,
see Fig.~\ref{fig:teaser}~(b).

Finally, we link the short tubelets with sufficient overlap
across temporally-overlapping short video segments.
If two boxes,
which are from the temporally-overlapping frame
(\ie\ the same frame)
of two neighboring short tubelets,
have sufficient spatial overlap,
the two corresponding short tubelets are linked together and merged.
We exploit the object linking
to improve the classification quality
by boosting the classification scores for positive detections
through aggregating the classification scores
of the linked tubelets.
As shown in Fig.~\ref{fig:teaser}~(c),
the detection results can be further improved
by considering long range temporal contexts.

Elaborate experiments are conducted
on the ImageNet VID dataset \cite{russakovsky2015imagenet}.
Our method obtains mAP $74.5\%$ training on the VID
and $80.6\%$ training on the mixture of VID and DET.
The results outperform both the static image detector
and the previous best performed methods.
In particular,
our method obtains $8.8\%$ absolute improvement
compared with the static image detector for fast moving objects.

\section{Related Work}
The task of object detection in both images \cite{felzenszwalb2010object,girshick2015fast,girshick2014rich,he2017mask,lin2017focal,liu2016ssd,viola2001rapid,redmon2016you} and videos \cite{han2016seq,feichtenhofer2017detect,kang2016object,kang2017object,kang2016t,zhu2017flow,zhu2016deep,wang2018fully,bertasius2018object,xiao2018video} has been widely studied in the literature.
We mainly review related works on video object detection and classify
them into three categories
by how they use the temporal contexts.

\vspace{.1cm}
\noindent\textbf{Feature Propagation w/o Object Linking.}
{In~\cite{zhu2017flow,wang2018fully,bertasius2018object,xiao2018video},
the features of the current frame are augmented by aggregating features propagated from neighboring frames.
The methods in \cite{zhu2017flow,wang2018fully} use the optical flows \cite{dosovitskiy2015flownet}
to spatially align features in different frames for feature propagation.
Bertasius \etal\ \cite{bertasius2018object} propagate features by using the deformable convolutional network \cite{dai2017deformable} across space and time.
Xiao and Lee \cite{xiao2018video} adapts the Conv-GRU \cite{ballas2015delving} to propagate features from neighboring frames.}
Feature propagation is also exploited in~\cite{zhu2016deep}
to speed up the object detection.
The authors propose to compute the feature maps
(using a very deep network with high computation cost)
for the key frames and propagate the features to non-key frames
by computing the optical flows using a shallow network which takes less time.
These methods are different from ours because they do not perform object linking.

\vspace{.1cm}
\noindent\textbf{Feature Propagation w/ Object Linking.}
The tubelet proposal network~\cite{kang2017object}
computes tubelets
by first generating static object proposals in the first frame
and then predicting their relative movements in following frames.
The features of the boxes in the tubelets
are propagated to each box for classification
by using a CNN-LSTM network.
{Apart from the feature propagation w/o object linking,
Wang \etal\ \cite{wang2018fully} also link object in neighboring frames for feature propagation.
More precisely, the relative movements in neighboring frames are predicted for each proposal in the current frame,
and the features of the boxes in neighboring frames are propagated to the corresponding box in the current frame by average pooling.}
Unlike these methods,
we link objects in the same frame
and propagate box scores instead of features across frames.
Besides, we directly generate the spatio-temporal cuboid proposals for video segments rather than per-frame proposals in \cite{kang2017object,wang2018fully}.

\vspace{.1cm}
\noindent\textbf{Score Propagation w/ Object Linking.}
The method in \cite{kang2016t,kang2016object} proposes two kinds of object linking.
The first one tracks the detected box in current frame
to its neighboring frames
to augment their original detections
for higher object recall.
The scores are also propagated to improve classification accuracy.
The linking is based on the mean optical flow vector within boxes.
The second one links objects into long tubelets using the tracking algorithm \cite{wang2015visual}
and then adopts a classifier
to aggregate the detection scores in the tubelets.
The Seq-NMS method~\cite{han2016seq} links objects
by checking the spatial overlap between boxes in neighboring frames
without considering the motion information
and then aggregates the scores of the linked objects for
the final score.
The method in \cite{feichtenhofer2017detect}
simultaneously predicts the object locations in two frames
and also the object movements from the preceding frame to the current frame.
Then they use the movements to link the detected objects into tubelets.
The object detection scores in the same tubelet is reweighed
by aggregating the scores in some manner
from the scores in that tubelet.

Our method belongs to the third category.
The main contribution of our work is that
we link objects in the same frame
instead of neighboring frames in previous methods \cite{han2016seq,kang2016object,kang2016t,feichtenhofer2017detect,kang2017object}.
In addition, to achieve our goal,
we develop a series of methods
such as cuboid proposal network
which have not been explored in previous methods.

\section{Method}

The task of video object detection
is to infer the locations and classes of the objects in each frame
of a video $\{\mathbf{I}^1, \mathbf{I}^2, \dots, \mathbf{I}^{N}\}$.
To obtain high quality object linking,
our method proposes to link objects in the same frame,
which can be used to improve the classification accuracy.

\begin{figure}[t]
\centering
\footnotesize
\includegraphics[width=0.75\linewidth]{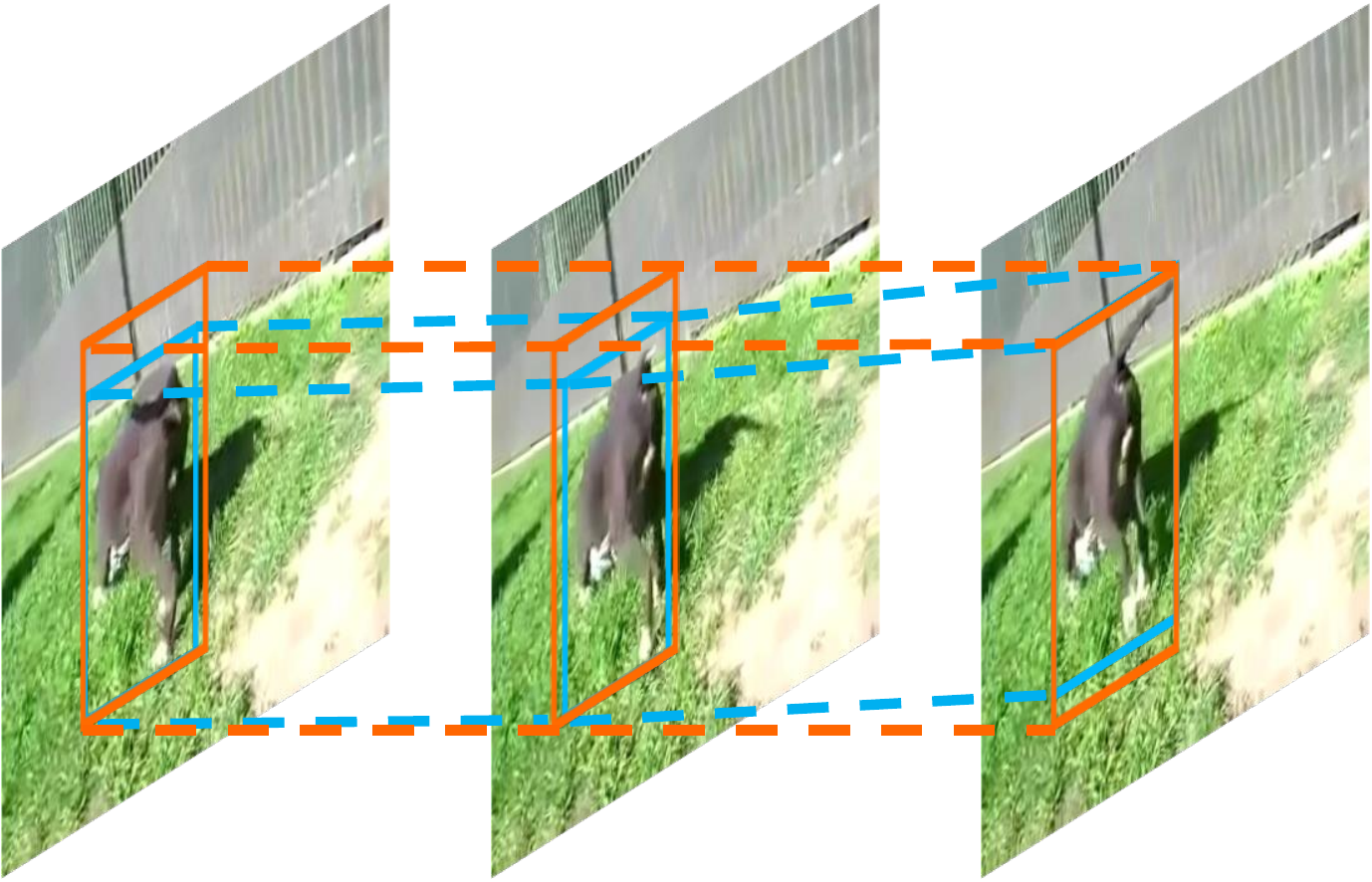}
\caption{
The orange cuboid, bounding the movement of the object,
is the target of the cuboid proposal stage.
The tubelet, composed of the blue object boxes in the video segment,
is the target of the short tubelet objection stage.
}
\label{fig:envelop}
%\vspace{-.4cm}
\end{figure}

\begin{figure}[t]
\centering
\footnotesize
\includegraphics[width=\linewidth]{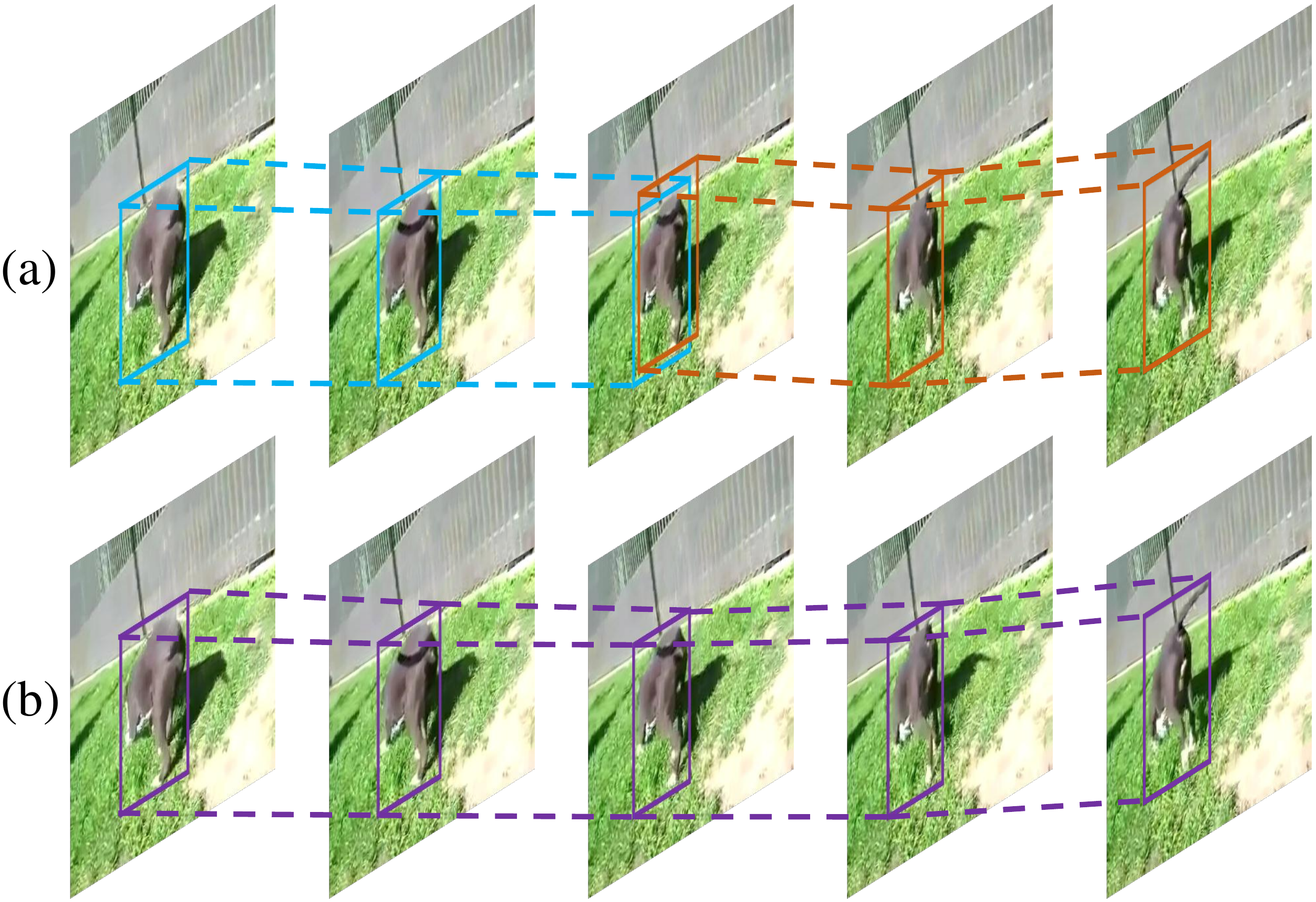}
\caption{
Illustration of short tubelet linking.
Boxes with the same color belong to the same tubelet.
The short tubelets (a) from two temporally-overlapping video segments are linked together to form new long tubelets (b).
}
\label{fig:linking}
%\vspace{-.4cm}
\end{figure}

Given a video divided into
a series of temporally-overlapping short video segments as the input,
our method consists of three stages:
(1)~Cuboid proposal generation for a short video segment.
This stage aims to generate a set of cuboids (containers)
which bound the same object across frames as shown in Fig.~\ref{fig:envelop}.
See Section~\ref{sec:cpg}.
(2)~Short tubelet detection for a short video segment.
For each cuboid proposal, the goal is to
regress and classify a short tubelet which is
a sequence of bounding boxes with each box
localizing the object in one frame.
The spatially-overlapping short tubelets
are removed by tubelet non-maximum suppression.
The short tubelet is a representation for linked objects across frames in a short video segment,
as illustrated in Fig.~\ref{fig:envelop}.
See Section~\ref{sec:std}.
(3)~Short tubelet linking for the whole video.
This stage, depicted in Fig.~\ref{fig:linking},
links the temporally-overlapping short tubelets
to link objects in the whole video,
and refines the classification scores of the linked tubelets.
See Section~\ref{sec:stl}.
The first two stages,
cuboid proposal generation
and short tubelet detection,
generate temporally-overlapping short tubelets,
and thus ensure that
we can link objects in the temporally-overlapping frame
(\ie\ the same frame)
in the short tubelet linking stage.

\subsection{Cuboid Proposal Generation}
\label{sec:cpg}

The ground truth bounding cuboid of the objects
in a short video segment,
containing $K$ frames
$\{\mathbf{I}^t, \mathbf{I}^{t+1}, \dots, \mathbf{I}^{t+K-1}\}$,
is defined as follows.
Let $\tilde{\mathbf{b}}$ be the $2D$ bounding box of
the tubelet $\tilde{\mathcal{T}} = (\tilde{\mathbf{b}}^{t}, \tilde{\mathbf{b}}^{t+1}, \dots,
\tilde{\mathbf{b}}^{t+K-1})$,
a series of all ground truth boxes
in the $K$ frames,
\begin{align}
\label{equ:gt_cpn}
\tilde{\mathbf{b}} = \operatorname{BoundingBox}(\tilde{\mathbf{b}}^{t}, \tilde{\mathbf{b}}^{t+1}, \dots,
\tilde{\mathbf{b}}^{t+K-1}).
\end{align}
Here, $\tilde{\mathbf{b}}^{\tau} = (x^{\tau}, y^{\tau}, w^{\tau}, h^{\tau})$ in frame ${\tau}$,
denoting the horizontal and vertical center coordinates
and its width and height,
is the ground truth box of frame $\tau$.
The bounding cuboid in our method is just a collection
of $K$ $\tilde{\mathbf{b}}$s: $\tilde{\mathbf{c}} = (\tilde{\mathbf{b}},
\tilde{\mathbf{b}}, \dots, \tilde{\mathbf{b}})$,
and thus simplified as a $2D$ box $\tilde{\mathbf{b}}$.
Fig.~\ref{fig:envelop}
provides the examples of the cuboid
and the tubelet in a short video segment.

We modify the region proposal network (RPN) method in Faster R-CNN~\cite{ren2015faster}
and introduce the cuboid proposal network (CPN) method
for computing cuboid proposals.
Unlike the conventional RPN
where the input is usually a single image,
our method takes the $K$ frames
as the input
to the CPN.
The output is a set of {$whk$} cuboid proposals,
regressed from a {$w \times h$} spatial grid,
where there are $k$ reference boxes at each location,
and each cuboid proposal is associated with an objectness score.

\subsection{Short Tubelet Detection}
\label{sec:std}

We use the $2D$ form of the cuboid proposal,
as the $2D$ box (region) proposal
for each frame in this segment,
which is classified and refined for each frame separately.

Considering a frame $\mathbf{I}^{\tau}$ in this segment,
we follow Fast R-CNN~\cite{girshick2015fast}
to refine the box and compute the classification score.
We start with a RoI pooling operation,
where the input is a $2D$ region proposal $\mathbf{b}$
and the response map of $\mathbf{I}^{\tau}$
obtained through a CNN.
The RoI pooling result is fed into
a classification layer,
outputting a $\{C+1\}$-dimensional classification score vector $\mathbf{y}^\tau$,
where $C$ is the number of categories and $1$ corresponds to the background,
as well as a regression layer,
from which the refined box
is obtained.

The resulting $K$ refined boxes for the $K$ frames
form the short tubelet detection result over this segment,
$\mathcal{T} = (\mathbf{b}^t, \mathbf{b}^{t+1}, \dots, \mathbf{b}^{t+K-1})$.
The classification score of this tubelet
is an aggregation of the scores over all the frames,
\begin{align}
\label{equ:score_aggregation}
\bar{\mathbf{y}} = \operatorname{aggregation}(\mathbf{y}^t, \mathbf{y}^{t+1}, \dots, \mathbf{y}^{t+K-1}),
\end{align}
where $\operatorname{aggregation}(\cdot)$ could be a $\operatorname{mean}$ operation.
We empirically find that
$\operatorname{Aggregation}(\cdot) = \frac{1}{2}(\operatorname{mean}(\cdot) + \operatorname{max}(\cdot))$
performs the best.

To remove spatial redundant short tubelets,
we extend the standard non-maximum suppression (NMS) algorithm
to a tubelet NMS (T-NMS) algorithm
to remove spatially-overlapping short tubelets in the same segment.
This strategy prevents tubelets from breaking by frame-wise NMS
which removes $2$D boxes for each frame independently.
The main point lies in
how to measure the spatial overlap
between two tubelets.
We define it on the base of
the overlap between the boxes
in the same frame.
Given two tubelets,
$\mathcal{T}_i = (\mathbf{b}^t_i, \mathbf{b}^{t+1}_i, \dots, \mathbf{b}^{t+K-1}_i)$
and $\mathcal{T}_j = (\mathbf{b}^t_j, \mathbf{b}^{t+1}_j, \dots, \mathbf{b}^{t+K-1}_j)$,
the spatial overlap is computed as
\begin{align}
\operatorname{overlap}(\mathcal{T}_i, \mathcal{T}_j)
=  \min_{\tau =t, t+1, \dots, t+K-1}\operatorname{IoU}(\mathbf{b}^{\tau}_i, \mathbf{b}^{\tau}_j),
\end{align}
where $\operatorname{IoU}(\mathbf{b}^{\tau}_i, \mathbf{b}^{\tau}_j)$
is the intersection over union between $\mathbf{b}^{\tau}_i$ and
$\mathbf{b}^{\tau}_j$ for frame $\tau$.
We choose this measurement
because two short tubelets are not the
same even if only one pair of corresponding boxes do not
have sufficient overlap.

\subsection{Short Tubelet Linking}
\label{sec:stl}
Our method divides a video
into a series of temporally-overlapping short video segments of length $K$
with stride $K-1$:
\begin{align}
& \mathcal{S}^1 = (\mathbf{I}^1, \mathbf{I}^2, \dots, \mathbf{I}^K), \\
& \mathcal{S}^2 = (\mathbf{I}^K, \mathbf{I}^{K+1}, \dots, \mathbf{I}^{2K-1}), \\
& \dots \dots \\
& \mathcal{S}^M = (\mathbf{I}^{(M-1)K-M+2}, \dots, \mathbf{I}^{MK-M+1}).
% & \mathcal{S}^M = (\mathbf{I}^{MK-M+1}, \mathbf{I}^{MK-M+2}, \dots, \mathbf{I}^{(M+1)K-M}).
\end{align}
Considering two temporally-overlapping short tubelets:
the $i$th tubelet from the $m$th segment
and the $i'$th tubelet from
the $(m+1)$th segment:
$$\mathcal{T}_i^m = (\mathbf{b}_i^{t_m}, \mathbf{b}_i^{t_m+1}, \dots, \mathbf{b}_i^{t_m+K-1}),$$
$$\mathcal{T}_{i'}^{m+1}
=(\mathbf{b}_{i'}^{t_m+K-1}, \mathbf{b}_{i'}^{t_m+K}, \dots, \mathbf{b}_{i'}^{t_m+2K-2}),$$
we link them if the spatial overlap between
$\mathbf{b}_i^{t_m+K-1}$ and $\mathbf{b}_{i'}^{t_m+K-1}$
from the temporally-overlapping frame (\ie\ the same frame)
is larger than a pre-defined threshold.

We perform a greedy short tubelet linking algorithm.
Initially, we put the short tubelets
from all short video segments
into a pool
and record the corresponding segment for each tubelet.
Our algorithm pops out the short tubelet $\mathcal{T}$ with the highest classification score from the pool.
We check the IoU of the boxes over the temporally-overlapping frame between $\mathcal{T}$ and its temporally-overlapping short tubelets.
If the IoU is larger than a threshold,
fixed as $0.4$ in our implementation,
we merge the two short tubelets into a single longer tubelet,
remove the box with the lower score for the overlapping frame,
update the classification score for the merged tubelet
according to Eq.~(\ref{equ:score_aggregation}) for better classification,
and record the corresponding segment (a combination of the corresponding two video segments).
We then push the merged tubelet into the pool.
This process is repeated
until no more tubelets can be merged.
Fig.~\ref{fig:linking} gives the examples of linking short tubelets to form long tubelets.

The tubelets remaining in the pool form
the video object detection results:
the score of the tubelet is assigned
to each box in the tubelet,
and the boxes from all the tubelets
associated with a frame
are regarded as the final detection boxes
for the corresponding frame.

\subsection{Implementation Details}
\label{sec:imple_details}

\noindent\textbf{Cuboid Proposal.}
The base network is ResNet-$101$ \cite{he2016deep} pre-trained on the ImageNet classification dataset \cite{russakovsky2015imagenet}:
we remove all layers after the \emph{Res5c} layer
and replace the convolutional layers in the fifth block
by dilated ones \cite{chen2017deeplab,yu2016multi} to reduce the stride from $32$ to $16$.
On the basis of the base network,
we add a convolutional layer with $512$ filters of $3\times3$,
and use two convolutional layers of $1\times1$ to
regress the offsets and predict the objectness scores for cuboid proposals.
The network is split into two sub-networks:
the first one has two residual blocks pass each frame separately to obtain frame-specific features
which are concatenated as input of the second sub-network with three residual blocks.

We use four anchor scales $64^2$, $128^2$, $256^2$, and $512^2$
with three aspect ratios $1$:$1$, $1$:$2$, and $2$:$1$,
resulting in $12$ anchors at each location in total.
The length $K$ of each video segment will be studied
in our experiments.
The loss function is the same as that in the standard RPN~\cite{ren2015faster}:
the cross-entropy loss for classification and the smoothed L1 loss for regression.
The training targets are the ground truth cuboids as defined in Eq.~(\ref{equ:gt_cpn}).
{The NMS threshold $0.7$ is chosen
and at most $300$ proposals are kept for the detection network training/testing.}
In the testing stage, if the number of frames in
the last segment is smaller than $K$,
we pad the segment by some frames copied from the last frame.

\vspace{.1cm}
\noindent\textbf{Short Tubelet Detection.}
The base network is the same as it for cuboid proposal.
We use RoI pooling to extract $7 \times 7$ response maps from the layer \emph{Res5c},
followed by two fully-connected + ReLU layers ($1024$ neurons).
{We use one fully-connected layer for classification
and another fully-connected layer for bounding box regression.}
Following the Fast R-CNN \cite{girshick2015fast},
we train the network
with online hard example mining~\cite{shrivastava2016training}.
The difference between our short tubelet detection training
and the Fast R-CNN training
is in the ground truth matching.
In particular,
we match a cuboid proposal to a ground truth box
if the IoU between the cuboid proposal
and a ground truth cuboid is larger than a threshold (typically $0.5$).
This is because the CPN is trained for cuboids,
which makes cuboid proposals hard to match ground truth boxes directly.
This matching strategy also ensures that a cuboid proposal
corresponds to the same object in different frames.
The training targets are still ground truth boxes
(rather than ground truth cuboids)
because we want to get accurate object locations in each frame.
During testing, the T-NMS threshold is set to $0.4$.

\vspace{.1cm}
\noindent\textbf{Training.}
We use SGD to train the cuboid proposal network
and the short tubelet detection network.
We initialize the weights of the newly added layers
by a zero-mean Gaussian distribution whose std is $0.01$.
Images are resized
to shorter side $600$ pixels
for both training and testing.
We set the mini-batch size to $8$,
the learning rate to $1\times10^{-3}$ for the first $40$K iterations
and $1\times 10^{-4}$ for the next $20$K iterations,
and the momentum to $0.9$.
We do not find the gain from sharing the base networks
for the cuboid proposal network
and the short tubelet detection network,
so we simply train them separately.
Our implementation is based on
the Caffe \cite{jia2014caffe} deep learning framework
on a TitanX (Pascal) GPU.

\subsection{Discussions}
\label{sec:discussions}

\noindent\textbf{Action Detection.}
The tasks of spatio-temporal action detection \cite{gkioxari2015finding}
and object detection in videos are similar to some extent.
The purpose of spatio-temporal action detection
is to localize and classify actions in each video frame.
Some solutions~\cite{gkioxari2015finding,hou2017tube,peng2016multi,saha2017amtnet,singh2017online,kalogeiton2017action} to action becomes similar to video object detection
and some of them can also be cast into the object/action linking framework.
For instance,
linking through neighboring frames, which is studied in video object detection~\cite{feichtenhofer2017detect,kang2016t,kang2017object,han2016seq},
is also explored
in~\cite{gkioxari2015finding,hou2017tube,peng2016multi,saha2017amtnet,singh2017online}.
We find that only the contemporary work~\cite{kalogeiton2017action} in action detection
adopts the scheme of linking through temporally-overlapping frames,
and its short tubelet detection scheme, similar to~\cite{kang2017object},
is different from our cuboid proposal based method.
It should be noted that although the solution frameworks
of the two problems are similar in high level,
the research focuses are different:
action detection is more about capturing the motion from the temporal signals
{and understanding an action from a single frame can be ambiguous (\eg\ sitting down or standing up) \cite{kalogeiton2017action},
whereas video object detection can be done in a single frame
and the temporal information is introduced to improve results in some frames of degraded image qualities.}
As validated in the later empirical results,
the state-of-art action detection method~\cite{kalogeiton2017action},
whose framework is similar to our method,
performs poor in video object detection.

\vspace{.1cm}
\noindent\textbf{Multi-object Issue.}
It is possible that one cuboid contains multiple objects,
because a cuboid tends to occupy a larger region than the object.
However, as observed in our experiments,
this problem has almost negligible influence on the detection performance.
This is because our overlapping-based short tubelet linking
can be accomplished when a video segment only has two frames.
In this case, each cuboid,
in most cases, contains only one object.
It is worth noting that it is not necessary
to use video segments longer than two frames,
because short two-frame segments already support
overlapping-based short tubelet linking.

\begin{figure}[t]
\centering
\footnotesize
\includegraphics[width=\linewidth]{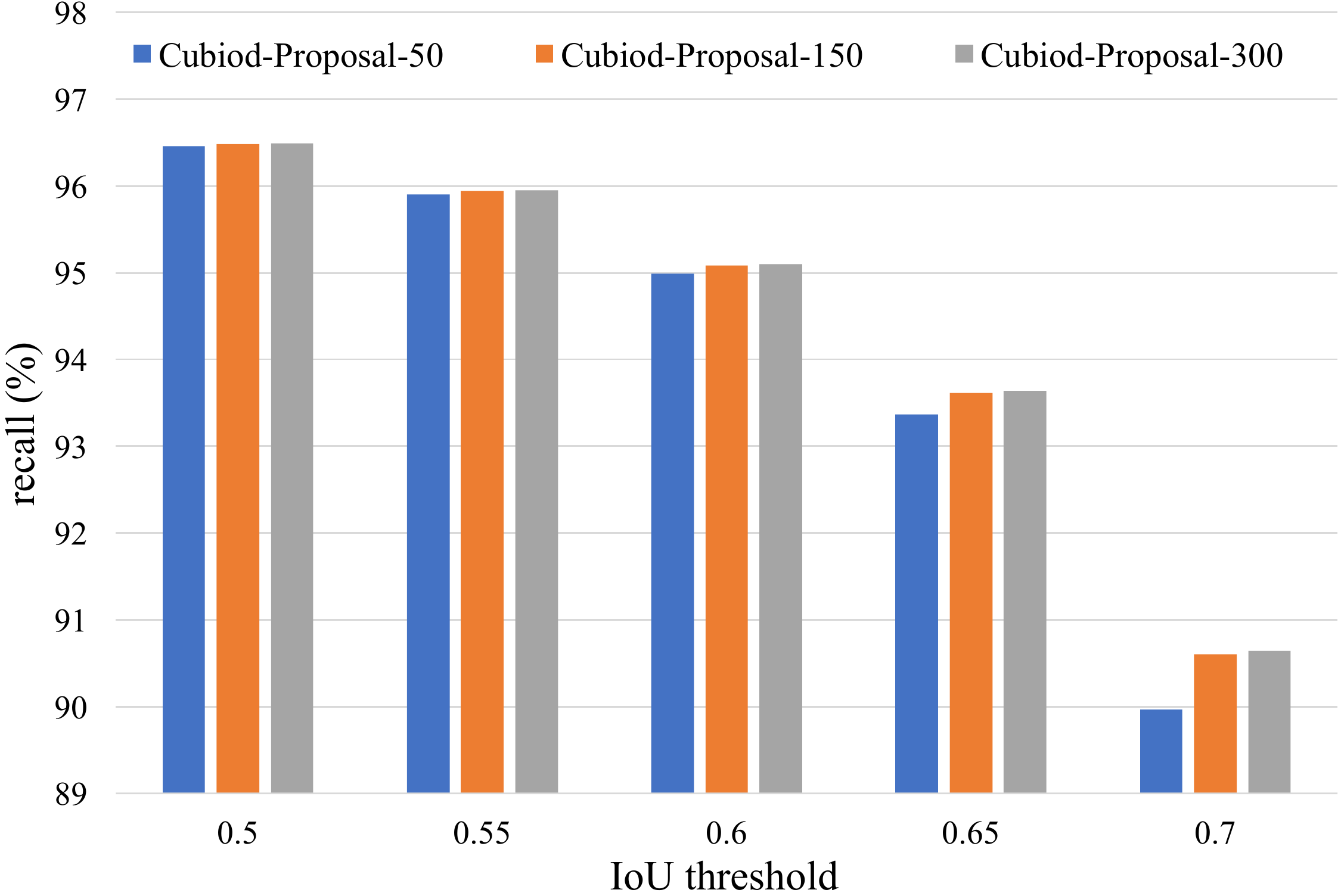}
\caption{Recall \vs\ IoU threshold on the VID validation set.
We show the results when keeping $50$, $150$, and $300$ cuboid proposals for IoU threshold $0.5$ to $0.7$.}
\label{fig:rpn_recall}
% \vspace{-.4cm}
\end{figure}

\vspace{.1cm}
\noindent\textbf{Boundary Issue.}
It is possible that in some frames an object may appear or disappear.
As a result, the boundary issue occurs in the short tubelet detection stage.
More precisely, in short tubelet detection,
$K$ successive frames share the same proposals and proposal classification scores.
Take $K=2$ as an example, there are two boundary frames, \ie\ the one frame before an object appears and the one frame after the object disappears.
The proposal classification scores of the two boundary frames will be enhanced according to Eq.~(\ref{equ:score_aggregation}),
which will result in false positives in these two frames.
This problem does not occur in the short tubelet linking stage
because we allow broken links for long range linking.
Actually in real applications,
the sequence where the object continuously appears is not short in most cases,
and then the two boundary frames will not affect the performance that much.
We investigate the VID dataset
and find that this boundary issue
only leads to small performance drop (up to $0.57\%$).

\section{Experiments}
\label{sec:experiment}

\subsection{Dataset and Evaluation Metric}
\label{sec:dataset}
We use the ImageNet VID dataset \cite{russakovsky2015imagenet}
which was introduced in the ILSVRC 2015 challenge.
The dataset contains $30$ object classes which cover different movement types and different levels of clutterness.
The dataset has $5354$ videos which are divided into training, validation, and testing subsets with $3862$, $555$, and $937$ videos, respectively.
Each video has about $300$ frames on average.
The dataset provides ground truth object locations, labels, and object identifications for each frame.
Since the annotations for the testing subset
has been reserved for the challenge
and the evaluation server has been closed,
we test on the validation subset as most of the other works.

We use the classical detection evaluation metric for the VID dataset,
\ie\ the Average Precision (AP) and mean of AP (mAP) over all classes,
following the previous works tested on VID \cite{feichtenhofer2017detect,kang2017object,kang2016t,kang2016object,zhu2017flow,zhu2016deep}.

\subsection{Ablation Studies}
\label{sec:ablation}

We first conduct detailed ablation experiments to study the effectiveness of different components in our method.
{For fair comparisons, the static image detector baseline mentioned below is a Faster R-CNN network \cite{ren2015faster}
that uses the same settings as we referred to in Section~\ref{sec:imple_details} except for treating all frames as static images without considering temporal information.}

\begin{figure}[t]
\centering
\footnotesize
\includegraphics[width=\linewidth]{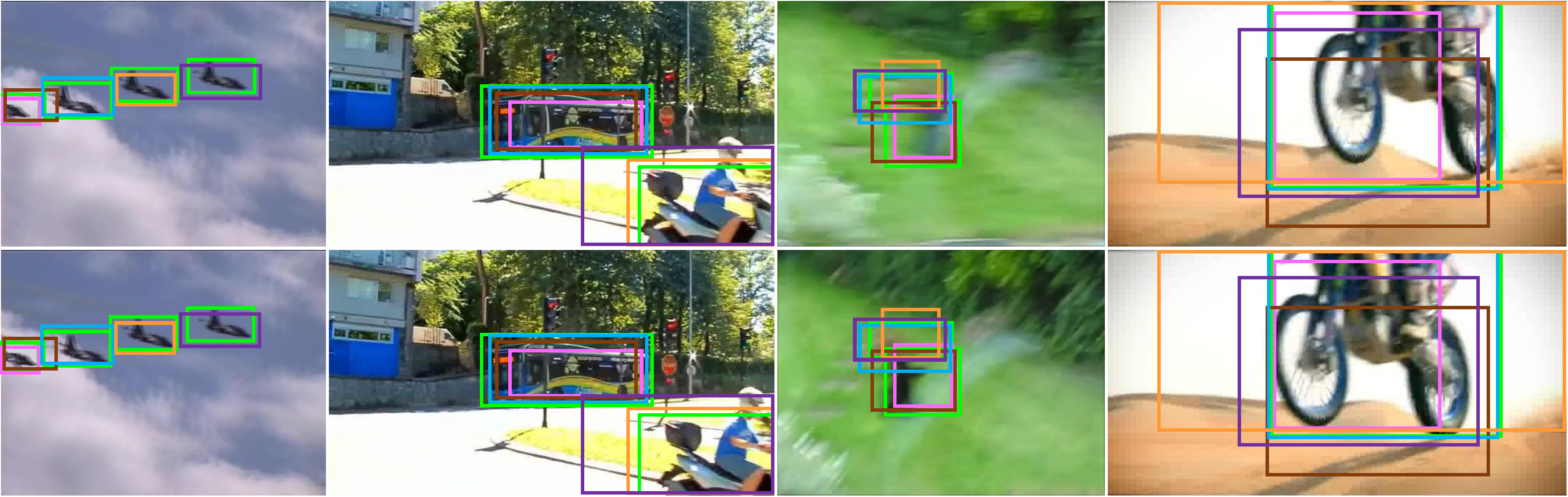}
\caption{Visualization of some cuboid proposal results.
Each column corresponds to a short video segment.
The green boxes are the ground truths and the rest are the cuboid proposals.
Boxes with the same color belong to the same cuboid proposals.
For each video segment, we only show five proposals with the highest objectness scores for simplicity.
}
\label{fig:rpn_qual}
% \vspace{-.4cm}
\end{figure}

\vspace{.1cm}
\noindent\textbf{Cuboid Proposal Recall.}
We first evaluate the recall of proposals by CPN.
To do this, we generate a collection of cuboid proposals for each video segment and compute their recall at different IoU thresholds ($0.5$ to $0.7$) with ground truth cuboids.
Fig.~\ref{fig:rpn_recall} shows the quantitative results on the validation set.
Firstly, we can see that keeping as few as $50$ proposals already gives reasonably good performance: more than $96.46\%$ of the ground truths are recalled for IoU $0.5$.
Secondly, increasing the number of proposals brings only marginal gains for lower IoU thresholds (\eg\ $0.5$) and gives larger gains for higher IoU thresholds (\eg\ $0.7$).
The results show that choosing $300$ proposals already achieves satisfactory recall.
Thus we only use $300$ proposals for following experiments.

We also show several qualitative results in Fig.~\ref{fig:rpn_qual}.
The green boxes are the ground truth cuboids and the rest are the proposals generated by CPN.
{In most cases, there is at least one proposal that has sufficient overlap with the ground truth cuboids,
which shows that the CPN can generate reliable cuboid proposals
and deals well with videos having single/multiple, small/large, fast/slow moving objects.}

\begin{figure}[t]
\centering
\footnotesize
\includegraphics[width=.94\linewidth]{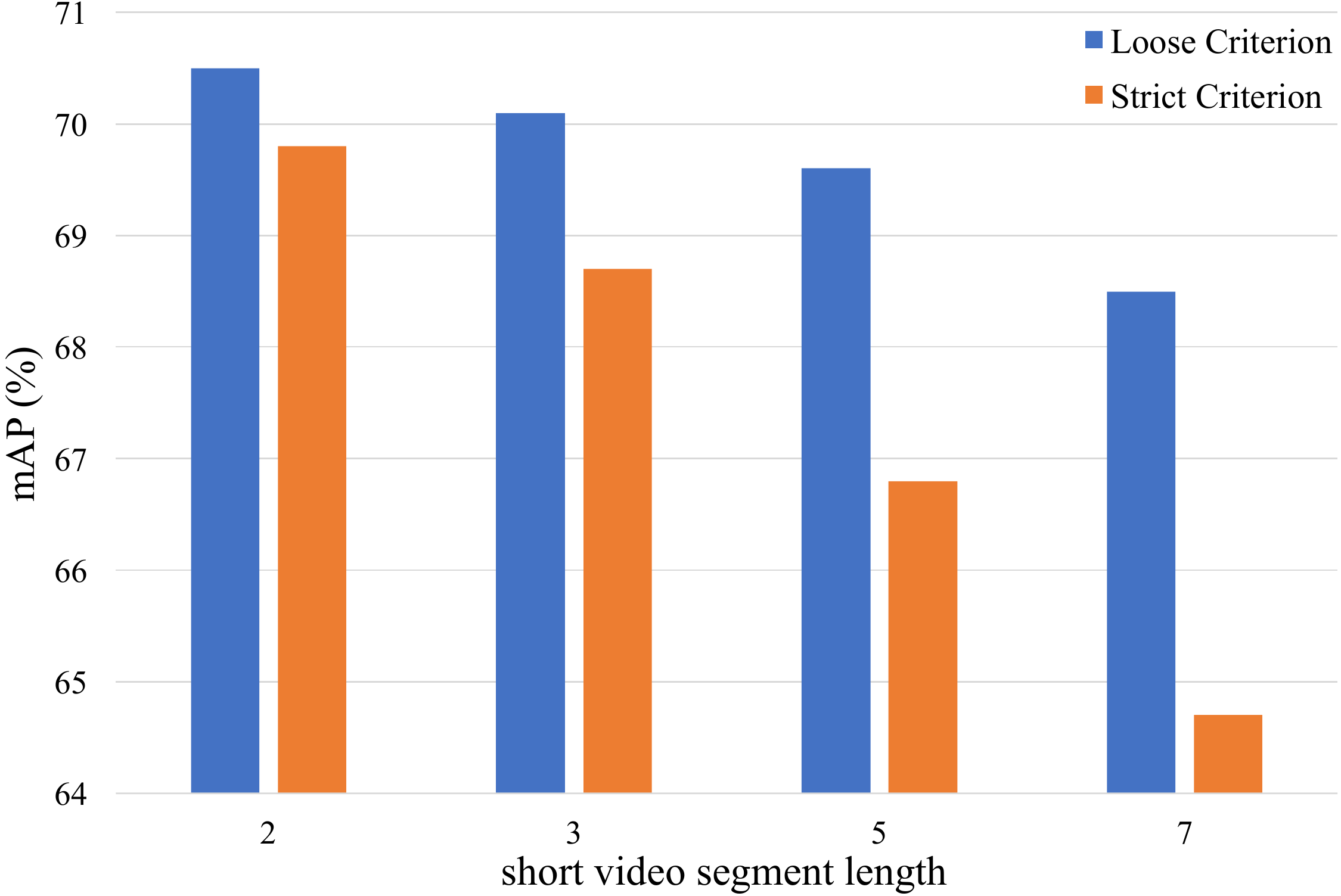}
\caption{Short tubelet detection results.
    The ``Loose Criterion'' is the standard detection mAP and the ``Strict Criterion'' checks whether the instance IDs of the boxes in a tubelet are the same.
}
\label{fig:shortlinking}
% \vspace{-.4cm}
\end{figure}

\begin{figure}[t]
\centering
\footnotesize
\includegraphics[width=.85\linewidth]{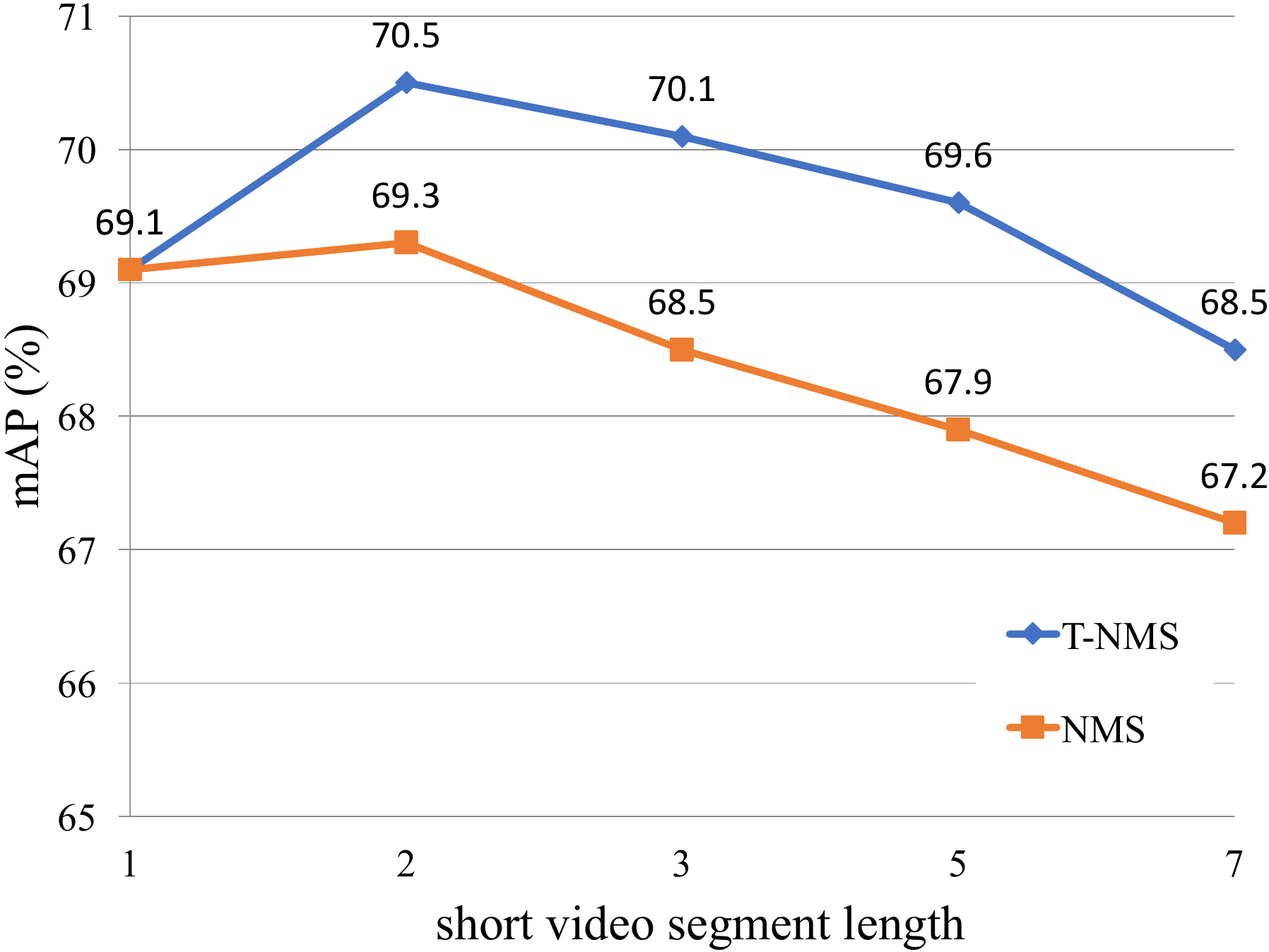}
\caption{Detection results for NMS/T-NMS and different short video segment lengths.
    Video segment length 1 means the static image detector.
}
\label{fig:ablation}
% \vspace{-.4cm}
\end{figure}

\vspace{.1cm}
\noindent\textbf{Short Tubelet Detection.}
\label{sec:linking}
We then investigate whether the boxes in the short tubelets for short video segments correspond to the same objects.
For a testing video, our method first generates a set of short tubelets.
Then if all boxes in the short tubelet localize object accurately and correspond to the same object,
the tubelet is classified as true positive,
and otherwise it is a false positive.
After that we compute the mAP.
It is obvious that this is a more strict evaluation criterion than the one used for video object detection.
Fig.~\ref{fig:shortlinking} shows the results.
We can see that using the strict evaluation protocol only slightly decreases the performance (\eg\ from $70.5\%$ to $69.8\%$ or from $70.1\%$ to $68.7\%$),
which justifies that the linking results of short tubelets are reasonably accurate.

\begin{table}[t]
\centering
\footnotesize
\caption{Detection results (mAP in $\%$) of different methods on the VID validation set which {has four subsets according to the object moving speed or occlusion}.
The relative gains over the static image detector baseline (a) are listed in the subscript.
The video segment length is set to 2 for (c-g).}
\label{table:ablation}
\resizebox{\linewidth}{!}{
\begin{tabular}{c|c|c|c|c|c|c|c}
   \hline
   Methods & (a) & (b) & (c) & (d) & {(e)} & {(f)} & (g)\\
   \hline
   {Union Proposal} &&&&&& {\checkmark} & \\
   Seq-NMS && \checkmark &&& {\checkmark} && \\
   CPN && &\checkmark & \checkmark & {\checkmark} && \checkmark \\
   T-NMS &&&& \checkmark && {\checkmark} & \checkmark\\
   Short Tubelet Linking &&&&&& {\checkmark} & \checkmark\\
   \hline
   mAP (\%) & $69.1$ & $70.9_{\uparrow 1.8}$ & $69.3_{\uparrow 0.2}$ & $70.5_{\uparrow 1.4}$ & {$72.1_{\uparrow 3.0}$} & {$72.6_{\uparrow 3.5}$} & $\foo{74.5_{\uparrow5.4}}$ \\
   \hline
   mAP (\%) (slow) & $76.8$ & $78.5_{\uparrow 1.7}$ & $76.7_{\downarrow 0.1}$ & $77.7_{\uparrow 0.9}$ & {$78.5_{\uparrow 1.7}$} & {$78.8_{\uparrow 2.0}$} & $\foo{80.4_{\uparrow 3.6}}$ \\
   \hline
   mAP (\%) (medium) & $68.5$ & $70.4_{\uparrow 1.9}$ & $68.9_{\uparrow 0.4}$ & $70.2_{\uparrow 1.7}$ & {$72.9_{\uparrow 4.4}$} & {$72.6_{\uparrow 4.1}$} & $\foo{74.7_{\uparrow 6.2}}$ \\
   \hline
   mAP (\%) (fast) & $47.2$ & $49.4_{\uparrow 2.2}$ & $47.7_{\uparrow 0.5}$ & $49.4_{\uparrow 2.2}$ & {$53.3_{\uparrow 6.1}$} & {$54.5_{\uparrow 7.3}$} & $\foo{56.0_{\uparrow 8.8}}$ \\
   \hline
   {mAP (\%) (occluded)} & {$65.4$} & {$67.6_{\uparrow 2.2}$} & {$65.2_{\downarrow 0.2}$} & {$66.4_{\uparrow 1.0}$} & {$68.3_{\uparrow 2.9}$} & {$68.7_{\uparrow 3.3}$} & {$70.4_{\uparrow 5.0}$} \\
   \hline
\end{tabular}
}
\end{table}

\vspace{.1cm}
\noindent\textbf{The Influence of Short Video Segment Length.}
\label{sec:length}
We discuss the influence of short video segment length.
From Fig.~\ref{fig:ablation},
we can see that using video segment lengths of $2$, $3$, and $5$ (with T-NMS) all improves over the static baseline.
The largest improvement ($1.4\%$ mAP) is obtained when the video segment length is $2$.
When the video segment length increases, the performance decreases.
In addition, as shown in Fig.~\ref{fig:shortlinking},
we can see that the short tubelet detection performance for long video segments is worse than short ones.
There are several reasons explaining this phenomenon.
First, longer segments are more probable to generate oversized proposals which have smaller overlap with the ground truth boxes in each frame.
Second, the oversized proposals are probable to overlap with the image regions of other objects,
causing more ambiguities for accurate localization and classification.
Due to the better object/tubelet detection results,
we set the video segment length to $2$ in the following if not specified.
In addition,
the video segment length larger than $1$ ensures that
we can link objects in the same frame.

\vspace{.1cm}
\noindent\textbf{NMS \vs\ T-NMS.}
\label{sec:nms}
We study the influence of NMS/T-NMS for object detection.
The NMS is implemented by removing boxes for each frame independently instead of removing short tubelets for video segments in the T-NMS.
Fig.~\ref{fig:ablation} and Table~\ref{table:ablation} show that T-NMS gives better performance than NMS,
which confirms that compared with NMS handling each frame independently,
simply considering short range temporal contexts contributes to better detection results.

\begin{table*}[t]
\centering
\footnotesize
\caption{Average precision (in $\%$) for different methods on the VID dataset.
$^{\star}$ indicates models trained on the mixture of VID and DET datasets.
}
\label{table:detection_results}

\resizebox{\linewidth}{!}{
\begin{tabular}{|@{}L{3.05cm}|*{15}{x}|x|}
\hline
Method & aero & antelope & bear & bike & bird & bus & car & cattle & dog & cat & elephant & fox & g$\_$panda & hamster & horse & \\
\hline
\hline
Kang \etal\ \cite{kang2017object} & 84.6 & 78.1 & 72.0 & 67.2 & 68.0 & 80.1 & 54.7 & 61.2 & 61.6 & 78.9 & 71.6 & 83.2 & 78.1 & 91.5 & 66.8 & \\
Kang \etal\ \cite{kang2016t}$^{\star}$ & 83.7 & 85.7 & 84.4 & 74.5 & 73.8 & 75.7 & 57.1 & 58.7 & 72.3 & 69.2 & 80.2 & 83.4 & 80.5 & 93.1 & \bf{84.2} & \\
Lee \etal\ \cite{lee2016multi}$^{\star}$ & 86.3 & 83.4 & 88.2 & \bf{78.9} & 65.9 & \bf{90.6} & \bf{66.3} & \bf{81.5} & 72.1 & 76.8 & 82.4 & 88.9 & \bf{91.3} & 89.3 & 66.5 & \\
Zhu \etal\ \cite{zhu2017flow}$^{\star}$ & - & - & - & - & - & - & - & - & - & - & - & - & - & - & - & \\
Feichtenhofer \etal\ \cite{feichtenhofer2017detect}$^{\star}$ & 90.2 & 82.3 & 87.9 & 70.1 & 73.2 & 87.7 & 57.0 & 80.6 & 77.3 & 82.6 & \bf{83.0} & \bf{97.8} & 85.8 & 96.6 & 82.1 & \\
{Wang \etal\ \cite{wang2018fully}$^{\star}$} & {88.7} & {\bf{88.4}} & {86.9} & {71.4} & {73.0} & {78.9} & {59.3} & {78.5} & {77.8} & {90.6} & {79.1} & {96.3} & {84.8} & {\bf{98.5}} & {77.4} &\\
{Bertasius \etal\ \cite{bertasius2018object}$^{\star}$} & {-} & {-} & {-} & {-} & {-} & {-} & {-} & {-} & {-} & {-} & {-} & {-} & {-} & {-} & {-} &\\
{Xiao and Lee \cite{xiao2018video}$^{\star}$} & {-} & {-} & {-} & {-} & {-} & {-} & {-} & {-} & {-} & {-} & {-} & {-} & {-} & {-} & {-} &\\
\hline
Ours & 89.9 & 77.8 & 81.7 & 71.6 & 71.9 & 85.3 & 60.0 & 69.9 & 69.4 & 85.7 & 79.9 & 90.2 & 83.4 & 93.5 & 67.3 & \\
Ours$^{\star}$ & \bf{90.5} & 80.1 & \bf{89.0} & 75.7 & \bf{75.5} & 83.5 & 64.0 & 71.4 & \bf{81.3} & \bf{92.3} & 80.0 & 96.1 & 87.6 & 97.8 & 77.5 & \\
\hline
\hline
Method & lion & lizard & monkey & mbike & rabbit & r$\_$panda & sheep & snake & squirrel & tiger & train & turtle & boat & whale & zebra & mAP \\
\hline\hline
Kang \etal\ \cite{kang2017object} & 21.6 & 74.4 & 36.6 & 76.3 & 51.4 & 70.6 & 64.2 & 61.2 & 42.3 & 84.8 & 78.1 & 77.2 & 61.5 & 66.9 & 88.5 & 68.4 \\
Kang \etal\ \cite{kang2016t}$^{\star}$ & 67.8 & 80.3 & 54.8 & 80.6 & 63.7 & 85.7 & 60.5 & 72.9 & 52.7 & 89.7 & 81.3 & 73.7 & 69.5 & 33.5 & 90.2 & 73.8 \\
Lee \etal\ \cite{lee2016multi}$^{\star}$ & 38.0 & 77.1 & 57.3 & \bf{88.8} & 78.2 & 77.7 & 40.6 & 50.3 & 44.3 & 91.8 & 78.2 & 75.1 & \bf{81.7} & 63.1 & 85.2 & 74.5 \\
Zhu \etal\ \cite{zhu2017flow}$^{\star}$ & - & - & - & - & - & - & - & - & - & - & - & - & - & - & - & 78.4 \\
Feichtenhofer \etal\ \cite{feichtenhofer2017detect}$^{\star}$ & 66.7 & 83.4 & \bf{57.6} & 86.7 & 74.2 & 91.6 & 59.7 & 76.4 & \bf{68.4} & \bf{92.6} & \bf{86.1} & \bf{84.3} & 69.7 & 66.3 & \bf{95.2} & 79.8 \\
{Wang \etal\ \cite{wang2018fully}$^{\star}$} & {\bf{75.5}} & {\bf{84.8}} & {55.1} & {85.8} & {76.7} & {\bf{95.3}} & {\bf{76.2}} & {75.7} & {59.0} & {91.5} & {81.7} & {84.2} & {69.1} & {72.9} & {94.6} & {80.3}\\
{Bertasius \etal\ \cite{bertasius2018object}$^{\star}$} & {-} & {-} & {-} & {-} & {-} & {-} & {-} & {-} & {-} & {-} & {-} & {-} & {-} & {-} & {-} & {80.4}\\
{Xiao and Lee \cite{xiao2018video}$^{\star}$} & {-} & {-} & {-} & {-} & {-} & {-} & {-} & {-} & {-} & {-} & {-} & {-} & {-} & {-} & {-} & {80.5}\\
\hline
Ours & 46.9 & 74.7 & 49.2 & 81.4 & 57.0 & 74.9 & 65.2 & 60.0 & 48.6 & 91.0 & 85.1 & 82.7 & 74.0 & 74.9 & 91.7 & 74.5 \\
Ours$^{\star}$ & 73.1 & 81.5 & 56.0 & 85.7 & \bf{79.9} & 87.0 & 68.8 & \bf{80.7} & 61.6 & 91.6 & 85.5 & 81.3 & 73.6 & \bf{77.4} & 91.9 & \bf{80.6} \\
\hline
\end{tabular}
}
\end{table*}

\vspace{.1cm}
\noindent\textbf{Short Tubelet Linking.}
\label{sec:longlink}
Here, we show the improvement by linking short tubelets.
We evaluate the performance on slow, medium, and fast ones
which are formed according to their speed as done in \cite{zhu2017flow}.
{We also evaluate the performance on occluded objects (\ie\ parts of objects are occluded), following \cite{wang2018fully} to select $87,195$ frames which have more than half occluded objects.}
As we can see in Table~\ref{table:ablation},
compared with the static baseline,
considering both short and long range temporal information boosts the performance.
When linking objects over the whole video to consider long range temporal context, there is significant improvements ($5.4\%$ to static and $4.0\%$ to without short tubelet linking).
Importantly, the performance gains are mainly from the faster objects ($6.2\%$ for medium and $8.8\%$ for fast).
It is natural that faster objects may have more variations,
thus detecting them depends more on temporal context.
{Our method also obtains $5.0\%$ performance gains for occluded objects,
which confirms that our method works well for occlusions.}
As short tubelet linking performs much better than others,
in the following we only report results by short tubelet linking.

\vspace{.1cm}
\noindent\textbf{Ours \vs\ Seq-NMS \cite{han2016seq}.}
We compare our results with
results by the linking in neighboring frames method Seq-NMS \cite{han2016seq}.
As shown in Table~\ref{table:ablation},
the Seq-NMS obtains better performance than the static baseline,
which also confirms the usefulness of temporal contexts.
However, the Seq-NMS performs much worse than our method.
In particular, the performance improvement for fast moving objects
by Seq-NMS is $2.2\%$,
whereas our method obtains $8.8\%$ improvement.
This is because the same object in neighboring frames
has different locations and appearances,
which influences the quality of object linking,
especially for fast moving objects.
Thus it is better to link objects in the same frame.

{We also combine our CPN and the Seq-NMS for object linking.
More precisely, we detect short tubelets from cuboid proposals,
and link short tubelets in neighboring frames similar to the Seq-NMS.
Results in Table~\ref{table:ablation} show that the CPN+Seq-NMS obtains better performance than the method that combines the static detector and the Seq-NMS.
This is because our short tubelet detection method can obtain better short tubelets than the Seq-NMS.
The CPN+Seq-NMS performs worse than our method,
which further demonstrates the effectiveness of our linking in the same frame strategy.}

{
\vspace{.1cm}
\noindent\textbf{CPN \vs\ Union Proposal.}
Here, we compare our CPN with a union proposal baseline.
Unlike our method that generates cuboid proposals by CPN,
the union proposal method first generates proposals for each frame separately using the static detector,
then links proposals in every two neighboring frames according to the proposal IoU,
and finally produces the union of linked proposals as cuboid proposals.
As shown in Table~\ref{table:ablation}~(f),
the detection performance by the union proposal method is worse than our method.
This is because the union proposal method generate cuboid proposals by linking boxes in neighboring frames similar to the Seq-NMS \cite{han2016seq},
which cannot obtain high quality cuboid proposals as ours.
The cuboid proposal recalls also demonstrate this:
95.1\% for IoU threshold 0.5 and 84.9\% for IoU threshold 0.7 (union proposal)
\vs\ 96.5\% for IoU threshold 0.5 and 90.6\% for IoU threshold 0.7 (CPN).
}

\vspace{.1cm}
\noindent\textbf{Comparison with the State-of-the-Art Action Detection Method \cite{kalogeiton2017action}.}
\label{sec:action}
Finally, we compare our result with the result from \cite{kalogeiton2017action}
which is the state-of-the-art solution in video action detection
and adopts the object/action linking framework
similar to our method and~\cite{kang2017object,feichtenhofer2017detect,kang2016t},
by deploying the method on VID.
The result by \cite{kalogeiton2017action} is $60.2\%$ mAP
which is much weaker than our $74.5\%$.
{The key point, making our approach perform better,
is that
the object detection schemes are different.
More specifically,
our method detects objects for each frame separately,
only using the information for the individual frame (with the same proposal for two neighboring frames).
\cite{kalogeiton2017action} localizes action boxes for different frames jointly,
thus resulting in poor localization quality.
More precisely,
\cite{kalogeiton2017action} stacks features from neighboring frames and uses the stacked features to predict the boxes of these neighboring frames jointly,
losing the explicit frame-wise information for predicting the corresponding action box.
This is also observed in \cite{zhu2017flow,wang2018fully}.

}

\subsection{Results}
\label{sec:det_results}

We compare our object detection results with the current state of the arts in Table~\ref{table:detection_results}.
First, when only training on the VID dataset, our method obtains the superior result $74.5\%$ mAP.
To pursue the state-of-the-art detection performance, we follow the previous methods \cite{feichtenhofer2017detect,kang2016t,zhu2017flow} to use the mixture of ImageNet VID and DET datasets for training the detection network,
and utilize the standard multi-scale training and testing \cite{he2015spatial}.
As we can see, comparing our $80.6\%$ with other methods using the same ResNet-101 network \cite{feichtenhofer2017detect,zhu2017flow,wang2018fully,xiao2018video}, our method obtains better performance,
which confirms the effectiveness of our linking strategy.
Importantly,
compared with \cite{feichtenhofer2017detect,zhu2017flow,kang2016t,kang2017object,wang2018fully,bertasius2018object,xiao2018video}
that link objects in neighboring frames,
our linking objects in the same frame strategy obtains better performance,
which demonstrates that our method can obtain higher quality object linking results.

In particular,
{the methods in \cite{zhu2017flow,bertasius2018object,xiao2018video,wang2018fully} combine feature propagation
and the score propagation method Seq-NMS \cite{han2016seq}
to obtain their results.}
Feichtenhofer \etal\ \cite{feichtenhofer2017detect} use more anchor scales to obtain better proposals and add a tracking loss to learn better features for performance improvement.
There are potential benefits from learning better features in the proposal and detection stages by incorporating other methods such as feature propagation and extra losses into our method.

\subsection{Qualitative Results}

\begin{figure*}[ht]
\centering
\footnotesize
\includegraphics[width=\linewidth]{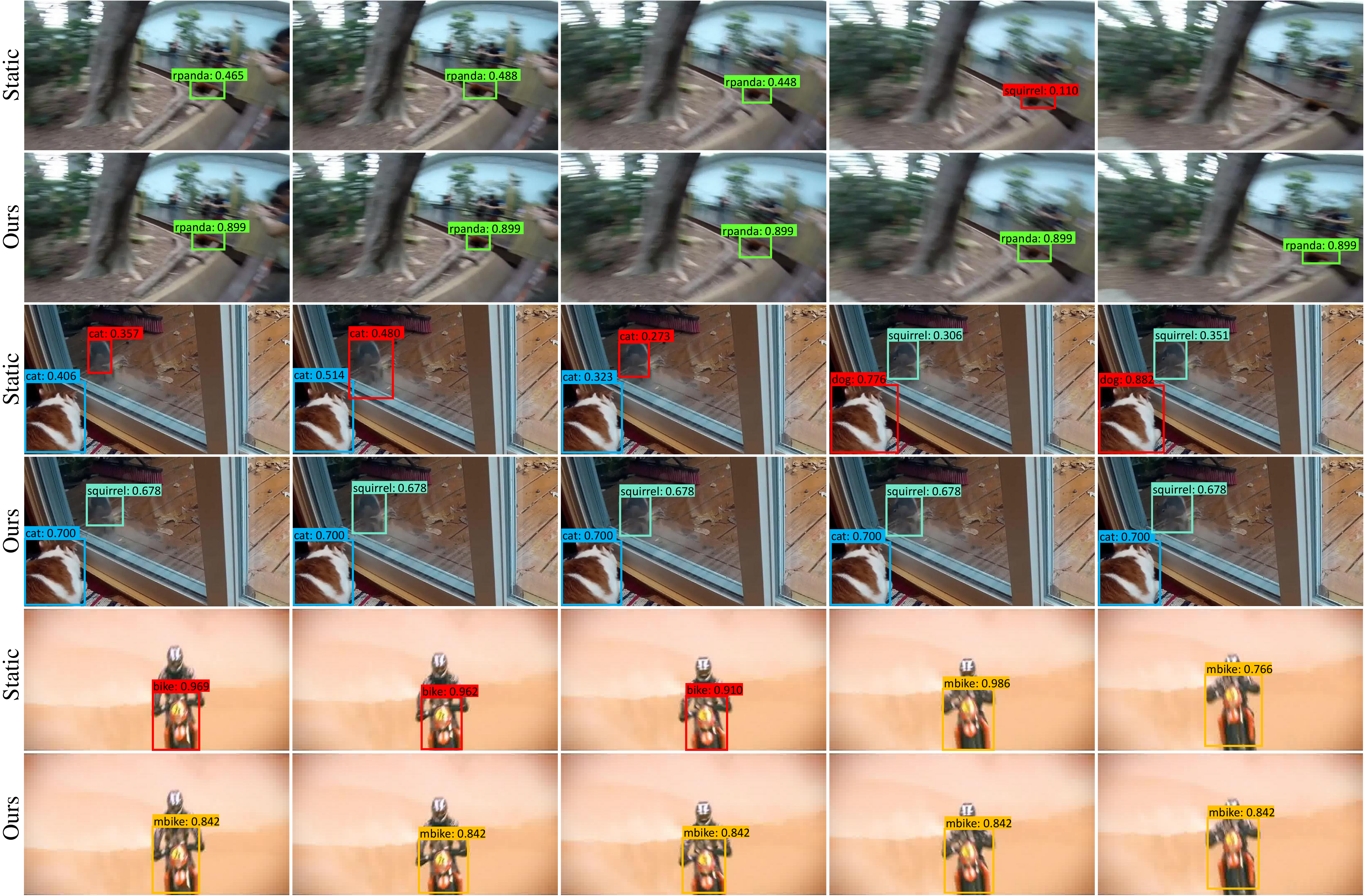}
\caption{Example detection results of the static detector and our method (only the top-scoring boxes around objects are shown).
Each row shows the results of sampled successive frames.
For the linking results, boxes with the same color belong to the same tubelet in the video.
Our method outperforms the static image detector when there are motion blurs, video defocus, and occlusions in the video.
Best viewed in color.}
\label{fig:detection_qual}
%\vspace{-.1cm}
\end{figure*}

Fig.~\ref{fig:detection_qual} visualizes several detection result comparisons between the static image detector and our method.
From the first two rows, we can see that the static method fails to detect the red-panda when there are severe motion blurs and occlusions.
This is reasonable because the appearance features have been severely degraded in this situation.
After applying the object linking and rescoring, our method successfully classifies the target in the challenging frames.
In addition, it is common that the static detectors may confuse with similar classes (\eg, bikes \vs\ motor-bikes, cats \vs\ dogs) especially when a frame has low image quality.
This problem can also be alleviated by rescoring the detections in the whole video because some frames have correct classifications and can propagate these scores to the challenging frames by object linking.

\begin{figure*}[ht]
\centering
\footnotesize
\includegraphics[width=\linewidth]{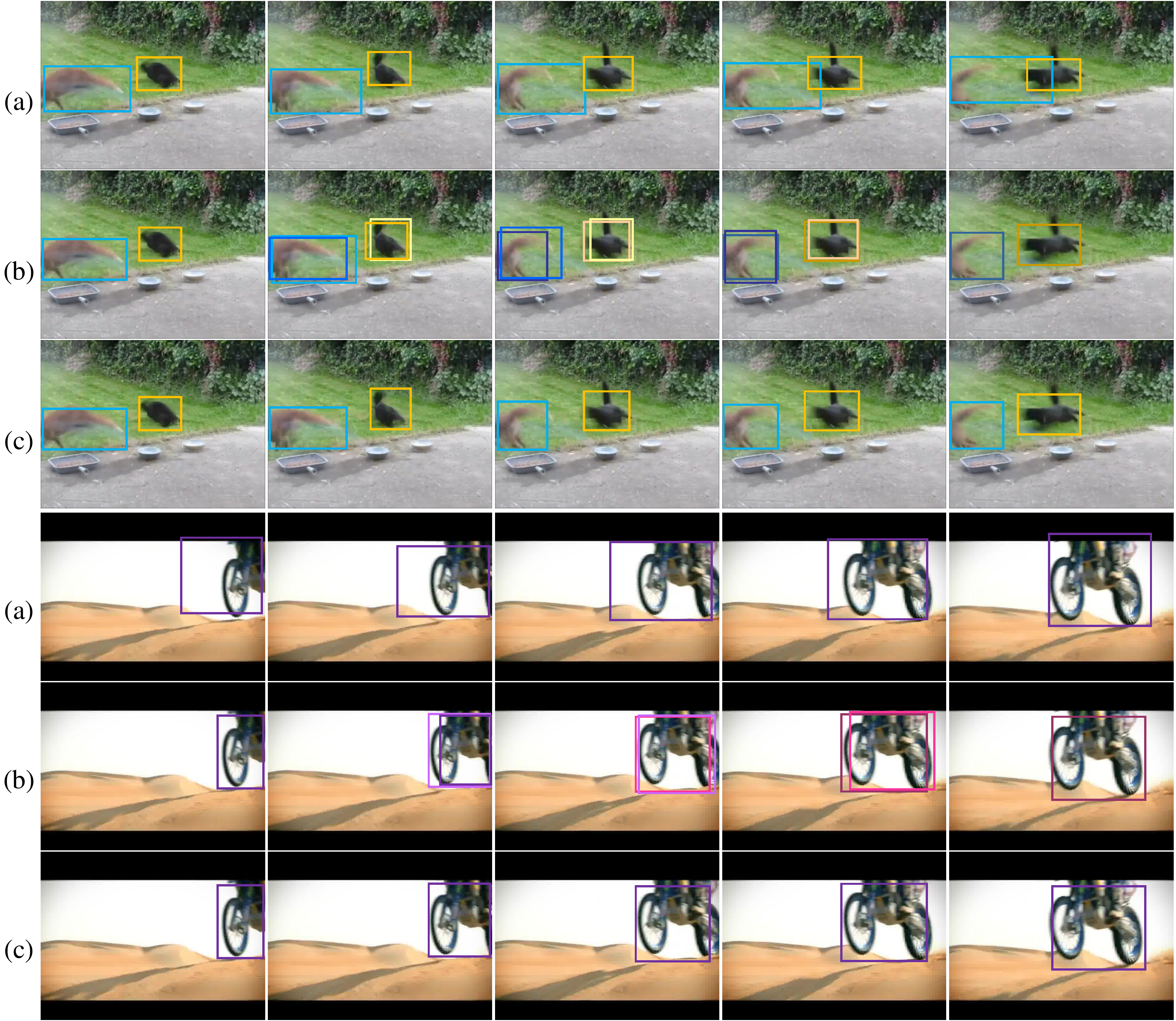}
\caption{Example object linking results of (a) static detector + Seq-NMS~\cite{han2016seq}, (b) short tubelet detection, and (c) short tubelet linking (only the top-scoring links around objects are shown).
Each row shows the results of sampled successive frames.
Boxes with the same color belong to the same tubelet.
The results of the baseline approach in (a) are poor in localization accuracy due to linking in neighboring frames.
In contrast, our results in (c) performs better
because CPN enables the later linking in the same frame scheme.
Best viewed in color.}
\label{fig:vis_link}
%\vspace{-.1cm}
\end{figure*}

To show that our cuboid proposal network (CPN) enabling linking in the same frame leads to better localization accuracy, Fig.~\ref{fig:vis_link} visualizes several object linking result comparisons between our method and the baseline approach of static detector + linking in neighboring frames over two examples.
For (a), we generate per-frame detection results using static image detector and then link
detection boxes in neighboring frames by Seq-NMS~\cite{han2016seq}.
For (b), the results are from our approach without later short tubelet linking.
One object in each frame in the second to fourth columns has two detected boxes.
For (c), the final results are from our approach with short tubelet linking.
From the two examples,
we can see that the localization accuracy in (a)
is poor because of linking in neighboring frames.
Here are the analyses.
In comparison to the per-frame region proposal network in static detector where the proposals across different frames are independent, the major benefits from CPN include:
(1) The proposals are associated.
A cuboid proposal consists of two per-frame proposals
that are thought to be about the same object,
and the resulting detected boxes (predicted for each frame separately) are also thought to be about the same object;
and (2) Two nearby cuboid proposals (as well as the resulting detected boxes),
\eg, one corresponds to the $(n-1)$-th and $n$-th frames,
and the other corresponds to the $n$-th and $(n+1)$-th frames, are spatially-overlapped 
in the $n$-th frame.
Consequently, our approach is able to link the detected boxes in the same frame.
The advantage in our linking in the same frame scheme is that
we do not need to care about the object movement.
In contrast, linking the detected boxes obtained from static detector
in neighboring frames
might suffer from the object movement
and harms the localization accuracy.

\subsection{Runtime}
For the case of two frame segments, our method takes $0.35$s per-frame for testing which is comparable to $0.30$s by the static baseline.
The small extra cost comes from the cuboid proposal generation procedure:
a small sub-network processing the two frames separately.
The extra time cost is small for the detection stage due to the shared convolutional feature map,
the computation time of T-NMS is almost the same as the NMS,
and the short tubelet linking is very efficient (about $10$ms per-frame).
The speed becomes even faster than the baseline when the short video segment length is larger than $2$ (\eg\ $0.27$s and $0.23$s for video segment length $3$ and $5$ respectively),
because the CPN generates cuboid proposals for all the frames in the segment by computing the features once.

\section{Conclusion}
\label{sec:conclusion}
In this paper,
we explore to link objects in the same frame
for high quality object linking
to improve the classification quality.
Our method has three main components to achieve our goal:
(1)~cuboid proposal network,
(2)~short tubelet detection,
and (3)~short tubelet linking.
Our method obtains the state-of-the-art video detection performance
on the VID dataset.

In the future,
we will extend our method
to handle the two main issues that our approach has:
the multi-object issue and the boundary issue.
The potential way for the first issue
is to generate multiple detection boxes from one proposal.
The potential way for the second issue
is to recheck the boxes in boundary frames separately.
In addition,
considering that feature propagation and score propagation are complementary to each other
as pointed out in \cite{zhu2017flow,wang2018fully,xiao2018video,bertasius2018object},
we will explore how to incorporate feature propagation and score propagation
for further performance improvement.

\section*{Acknowledgements}
This work was supported by National Natural Science Foundation of China
(No. 61733007, No. 61572207, No. 61876212),
Hubei Scientific and Technical Innovation Key Project, the Program for HUST Academic Frontier Youth Team, and CCF-Tencent Open Research Fund.

% or


% \ifCLASSOPTIONcompsoc
%   \section*{Acknowledgments}
% \else
% %   \section*{Acknowledgment}
% \fi


\ifCLASSOPTIONcaptionsoff
  \newpage
\fi



\bibliographystyle{IEEEtran}
\bibliography{egbib}

\begin{thebibliography}{10}
\providecommand{\url}[1]{#1}
\csname url@samestyle\endcsname
\providecommand{\newblock}{\relax}
\providecommand{\bibinfo}[2]{#2}
\providecommand{\BIBentrySTDinterwordspacing}{\spaceskip=0pt\relax}
\providecommand{\BIBentryALTinterwordstretchfactor}{4}
\providecommand{\BIBentryALTinterwordspacing}{\spaceskip=\fontdimen2\font plus
\BIBentryALTinterwordstretchfactor\fontdimen3\font minus
  \fontdimen4\font\relax}
\providecommand{\BIBforeignlanguage}[2]{{%
\expandafter\ifx\csname l@#1\endcsname\relax
\typeout{** WARNING: IEEEtran.bst: No hyphenation pattern has been}%
\typeout{** loaded for the language `#1'. Using the pattern for}%
\typeout{** the default language instead.}%
\else
\language=\csname l@#1\endcsname
\fi
#2}}
\providecommand{\BIBdecl}{\relax}
\BIBdecl

\bibitem{girshick2015fast}
R.~Girshick, ``Fast {R-CNN},'' in \emph{ICCV}, 2015, pp. 1440--1448.

\bibitem{girshick2014rich}
R.~Girshick, J.~Donahue, T.~Darrell, and J.~Malik, ``Rich feature hierarchies
  for accurate object detection and semantic segmentation,'' in \emph{CVPR},
  2014, pp. 580--587.

\bibitem{liu2016ssd}
W.~Liu, D.~Anguelov, D.~Erhan, C.~Szegedy, S.~Reed, C.-Y. Fu, and A.~C. Berg,
  ``{SSD}: Single shot multibox detector,'' in \emph{ECCV}, 2016, pp. 21--37.

\bibitem{redmon2016you}
J.~Redmon, S.~Divvala, R.~Girshick, and A.~Farhadi, ``You only look once:
  Unified, real-time object detection,'' in \emph{CVPR}, 2016, pp. 779--788.

\bibitem{ren2015faster}
S.~Ren, K.~He, R.~Girshick, and J.~Sun, ``Faster {R-CNN}: Towards real-time
  object detection with region proposal networks,'' in \emph{NIPS}, 2015, pp.
  91--99.

\bibitem{tang2017multiple}
P.~Tang, X.~Wang, X.~Bai, and W.~Liu, ``Multiple instance detection network
  with online instance classifier refinement,'' in \emph{CVPR}, 2017, pp.
  2843--2851.

\bibitem{zhang2018single}
Z.~Zhang, S.~Qiao, C.~Xie, W.~Shen, B.~Wang, and A.~L. Yuille, ``Single-shot
  object detection with enriched semantics,'' in \emph{CVPR}, 2018, pp.
  5813--5821.

\bibitem{he2016deep}
K.~He, X.~Zhang, S.~Ren, and J.~Sun, ``Deep residual learning for image
  recognition,'' in \emph{CVPR}, 2016, pp. 770--778.

\bibitem{krizhevsky2012imagenet}
A.~Krizhevsky, I.~Sutskever, and G.~E. Hinton, ``{ImageNet} classification with
  deep convolutional neural networks,'' in \emph{NIPS}, 2012, pp. 1097--1105.

\bibitem{lecun1998gradient}
Y.~LeCun, L.~Bottou, Y.~Bengio, and P.~Haffner, ``Gradient-based learning
  applied to document recognition,'' \emph{Proceedings of the IEEE}, vol.~86,
  no.~11, pp. 2278--2324, 1998.

\bibitem{simonyan2015very}
K.~Simonyan and A.~Zisserman, ``Very deep convolutional networks for
  large-scale image recognition,'' in \emph{ICLR}, 2015.

\bibitem{han2016seq}
W.~Han, P.~Khorrami, T.~L. Paine, P.~Ramachandran, M.~Babaeizadeh, H.~Shi,
  J.~Li, S.~Yan, and T.~S. Huang, ``Seq-{NMS} for video object detection,''
  \emph{arXiv preprint arXiv:1602.08465}, 2016.

\bibitem{feichtenhofer2017detect}
C.~Feichtenhofer, A.~Pinz, and A.~Zisserman, ``Detect to track and track to
  detect,'' in \emph{ICCV}, 2017, pp. 3038--3046.

\bibitem{kang2016object}
K.~Kang, W.~Ouyang, H.~Li, and X.~Wang, ``Object detection from video tubelets
  with convolutional neural networks,'' in \emph{CVPR}, 2016, pp. 817--825.

\bibitem{kang2016t}
K.~Kang, H.~Li, J.~Yan, X.~Zeng, B.~Yang, T.~Xiao, C.~Zhang, Z.~Wang, R.~Wang,
  X.~Wang \emph{et~al.}, ``{T-CNN}: Tubelets with convolutional neural networks
  for object detection from videos,'' \emph{arXiv preprint arXiv:1604.02532},
  2016.

\bibitem{kang2017object}
K.~Kang, H.~Li, T.~Xiao, W.~Ouyang, J.~Yan, X.~Liu, and X.~Wang, ``Object
  detection in videos with tubelet proposal networks,'' in \emph{CVPR}, 2017,
  pp. 727--735.

\bibitem{zhu2016deep}
X.~Zhu, Y.~Xiong, J.~Dai, L.~Yuan, and Y.~Wei, ``Deep feature flow for video
  recognition,'' in \emph{CVPR}, 2017, pp. 2349--2358.

\bibitem{zhu2017flow}
X.~Zhu, Y.~Wang, J.~Dai, L.~Yuan, and Y.~Wei, ``Flow-guided feature aggregation
  for video object detection,'' in \emph{ICCV}, 2017, pp. 408--417.

\bibitem{wang2015visual}
L.~Wang, W.~Ouyang, X.~Wang, and H.~Lu, ``Visual tracking with fully
  convolutional networks,'' in \emph{ICCV}, 2015, pp. 3119--3127.

\bibitem{bertasius2018object}
G.~Bertasius, L.~Torresani, and J.~Shi, ``Object detection in video with
  spatiotemporal sampling networks,'' in \emph{ECCV}, 2018, pp. 331--346.

\bibitem{xiao2018video}
F.~Xiao and Y.~Jae~Lee, ``Video object detection with an aligned
  spatial-temporal memory,'' in \emph{ECCV}, 2018, pp. 485--501.

\bibitem{russakovsky2015imagenet}
O.~Russakovsky, J.~Deng, H.~Su, J.~Krause, S.~Satheesh, S.~Ma, Z.~Huang,
  A.~Karpathy, A.~Khosla, M.~Bernstein \emph{et~al.}, ``{ImageNet} large scale
  visual recognition challenge,'' \emph{IJCV}, vol. 115, no.~3, pp. 211--252,
  2015.

\bibitem{felzenszwalb2010object}
P.~F. Felzenszwalb, R.~B. Girshick, D.~McAllester, and D.~Ramanan, ``Object
  detection with discriminatively trained part-based models,'' \emph{TPAMI},
  vol.~32, no.~9, pp. 1627--1645, 2010.

\bibitem{he2017mask}
K.~He, G.~Gkioxari, P.~Dollar, and R.~Girshick, ``Mask {R-CNN},'' in
  \emph{ICCV}, 2017, pp. 2961--2969.

\bibitem{lin2017focal}
T.-Y. Lin, P.~Goyal, R.~Girshick, K.~He, and P.~Dollar, ``Focal loss for dense
  object detection,'' in \emph{ICCV}, 2017, pp. 2980--2988.

\bibitem{viola2001rapid}
P.~Viola and M.~Jones, ``Rapid object detection using a boosted cascade of
  simple features,'' in \emph{CVPR}, 2001.

\bibitem{wang2018fully}
S.~Wang, Y.~Zhou, J.~Yan, and Z.~Deng, ``Fully motion-aware network for video
  object detection,'' in \emph{ECCV}, 2018, pp. 542--557.

\bibitem{dosovitskiy2015flownet}
A.~Dosovitskiy, P.~Fischer, E.~Ilg, P.~Hausser, C.~Hazirbas, V.~Golkov,
  P.~van~der Smagt, D.~Cremers, and T.~Brox, ``{FlowNet}: Learning optical flow
  with convolutional networks,'' in \emph{ICCV}, 2015, pp. 2758--2766.

\bibitem{dai2017deformable}
J.~Dai, H.~Qi, Y.~Xiong, Y.~Li, G.~Zhang, H.~Hu, and Y.~Wei, ``Deformable
  convolutional networks,'' in \emph{ICCV}, 2017, pp. 764--773.

\bibitem{ballas2015delving}
N.~Ballas, L.~Yao, C.~Pal, and A.~Courville, ``Delving deeper into
  convolutional networks for learning video representations,'' \emph{arXiv
  preprint arXiv:1511.06432}, 2015.

\bibitem{chen2017deeplab}
L.-C. Chen, G.~Papandreou, I.~Kokkinos, K.~Murphy, and A.~L. Yuille,
  ``{DeepLab}: Semantic image segmentation with deep convolutional nets, atrous
  convolution, and fully connected {CRFs},'' \emph{TPAMI}, 2017.

\bibitem{yu2016multi}
F.~Yu and V.~Koltun, ``Multi-scale context aggregation by dilated
  convolutions,'' in \emph{ICLR}, 2016.

\bibitem{shrivastava2016training}
A.~Shrivastava, A.~Gupta, and R.~Girshick, ``Training region-based object
  detectors with online hard example mining,'' in \emph{CVPR}, 2016, pp.
  761--769.

\bibitem{jia2014caffe}
Y.~Jia, E.~Shelhamer, J.~Donahue, S.~Karayev, J.~Long, R.~Girshick,
  S.~Guadarrama, and T.~Darrell, ``Caffe: Convolutional architecture for fast
  feature embedding,'' in \emph{ACM MM}, 2014, pp. 675--678.

\bibitem{gkioxari2015finding}
G.~Gkioxari and J.~Malik, ``Finding action tubes,'' in \emph{CVPR}, 2015, pp.
  759--768.

\bibitem{hou2017tube}
R.~Hou, C.~Chen, and M.~Shah, ``Tube convolutional neural network ({T-CNN}) for
  action detection in videos,'' in \emph{ICCV}, 2017, pp. 5822--5831.

\bibitem{peng2016multi}
X.~Peng and C.~Schmid, ``Multi-region two-stream {R-CNN} for action
  detection,'' in \emph{ECCV}, 2016, pp. 744--759.

\bibitem{saha2017amtnet}
S.~Saha, G.~Singh, and F.~Cuzzolin, ``{AMTnet}: Action-micro-tube regression by
  end-to-end trainable deep architecture,'' in \emph{ICCV}, 2017, pp.
  4414--4423.

\bibitem{singh2017online}
G.~Singh, S.~Saha, M.~Sapienza, P.~Torr, and F.~Cuzzolin, ``Online real-time
  multiple spatiotemporal action localisation and prediction,'' in \emph{ICCV},
  2017, pp. 3637--3646.

\bibitem{kalogeiton2017action}
V.~Kalogeiton, P.~Weinzaepfel, V.~Ferrari, and C.~Schmid, ``Action tubelet
  detector for spatio-temporal action localization,'' in \emph{ICCV}, 2017, pp.
  4405--4413.

\bibitem{lee2016multi}
B.~Lee, E.~Erdenee, S.~Jin, M.~Y. Nam, Y.~G. Jung, and P.~K. Rhee,
  ``Multi-class multi-object tracking using changing point detection,'' in
  \emph{ECCV}, 2016, pp. 68--83.

\bibitem{he2015spatial}
K.~He, X.~Zhang, S.~Ren, and J.~Sun, ``Spatial pyramid pooling in deep
  convolutional networks for visual recognition,'' \emph{TPAMI}, vol.~37,
  no.~9, pp. 1904--1916, 2015.

\end{thebibliography}
\end{document}